%% file: template.tex
\documentclass{Interspeech2024}
\usepackage{multicol}
\usepackage{multirow}
\usepackage{threeparttable,booktabs}
\usepackage{booktabs}
\usepackage{cite}
\usepackage{bbm}
\usepackage{amsmath}
\usepackage{dsfont}
\usepackage{algorithm}
\usepackage{arydshln}
\usepackage{resizegather}
\usepackage{threeparttable,booktabs}
\newcommand{\argmin}{\operatornamewithlimits{argmin}}

\usepackage{graphicx}               % Necessary to use \scalebox
\usepackage{amsmath,amssymb}
\usepackage[noend]{algpseudocode}
\usepackage{amssymb}
\usepackage{pifont}% http://ctan.org/pkg/pifont
\newcommand{\cmark}{\ding{51}}%
\newcommand{\xmark}{\ding{55}}%

\usepackage[utf8]{inputenc}
\usepackage[english]{babel}
\usepackage{siunitx}
\usepackage{multirow}
\usepackage{caption}

\usepackage{tabstackengine}
\TABstackMath
\setstackgap{S}{2pt}

\def\ie{\emph{i.e.,}}

\interspeechcameraready

\title{MSRS: Training Multimodal Speech Recognition Models from Scratch with Sparse Mask Optimization}

\name[affiliation={1}]{Adriana}{Fernandez-Lopez}
\name[affiliation={1}]{Honglie}{Chen}
\name[affiliation={1,2}]{Pingchuan}{Ma}
\name[affiliation={3}]{Lu}{Yin}
\name[affiliation={4}]{Qiao}{Xiao}
\name[affiliation={1,2}]{Stavros}{Petridis}
\name[affiliation={5}]{Shiwei}{Liu}
\name[affiliation={1,2}]{Maja}{Pantic}

\address{
  $^1$Meta AI, UK \quad
  $^2$Imperial College London, UK \quad
  $^3$University of Surrey \\
  $^4$Eindhoven University of Technology \quad
  $^5$University of Oxford 
  }
\email{\{afernandezlopez,hongliechen,pingchuanma\}@meta.com}


\begin{document}

\maketitle

\begin{abstract}
\vspace{-0.1cm}
Pre-trained models have been a foundational approach in speech recognition, albeit with associated additional costs. In this study, we propose a regularization technique that facilitates the training of visual and audio-visual speech recognition models (VSR and AVSR) from scratch. This approach, abbreviated as \textbf{MSRS} (Multimodal Speech Recognition from Scratch), introduces a sparse regularization that rapidly learns sparse structures within the dense model at the very beginning of training, which receives healthier gradient flow than the dense equivalent. Once the sparse mask stabilizes, our method allows transitioning to a dense model or keeping a sparse model by updating non-zero values. MSRS achieves competitive results in VSR and AVSR with 21.1\% and 0.9\% WER on the LRS3 benchmark, while reducing training time by at least 2$\times$. We explore other sparse approaches and show that only MSRS enables training from scratch by implicitly masking the weights affected by vanishing gradients.  

\end{abstract}
\noindent speech recognition, sparse mask optimization, sparse regularization, sparse networks.

\input{sections/intro}
\input{sections/method}

\input{sections/experimental_setup}
\input{sections/results}
\input{sections/conclusions}

\section{Acknowledgements}
    Only non-Meta authors conducted any of the dataset pre-processing (no dataset pre-processing took place on Meta’s servers or facilities). Shiwei Liu is supported by the Royal Society with the Newton International Fellowship.

\bibliographystyle{IEEEtran}
\bibliography{mybib}

\end{document}

%% file: sections/intro.tex
\section{Introduction}
\vspace{-0.1cm}
\label{sec:intro}

Automatic Speech Recognition (ASR), Visual Speech Recognition (VSR) and Audio-Visual Speech Recognition (AVSR) systems use various input sources (audio, video, and audio-visual) to map spoken language into written text. The success of these systems is due to advanced deep neural architectures \cite{ma2023auto, shi2022learning, haliassos2022jointly, afouras2018deep} and large-scale audiovisual datasets \cite{son2017lip,afouras2018deep}, where performance usually scales with both model size and training data. However, training massive deep networks is challenging and comes with well-known hurdles such as overfitting or vanishing/exploding gradients. Regularization strategies such as dropout, batch normalization, warm-up, gradient clipping, and data augmentation have been adopted but are not always sufficient for training deep VSR/AVSR models from scratch \cite{shi2022learning, haliassos2022jointly}. Current VSR/AVSR methods typically follow a two-stage curriculum learning approach \cite{bengio2009curriculum}, involving initial pretraining on a small subset of data (1$-$5 cycles \cite{ma2021towards, ma2022visual}), followed by continued training on a large-scale dataset \cite{ma2021towards, ma2021end}. Another line of work achieves a better convergence based on sub-words learning schemes, where training samples are cropped out with frame-word boundaries~\cite{prajwal2022sub}. In some cases, auxiliary tasks are also incorporated \cite{ma2022visual, shi2022learning}. Despite their success in training VSR/AVSR models, pretraining can be resource-intensive as the model and dataset size grows \cite{haliassos2022jointly,djilali2024vsr}. 

\input{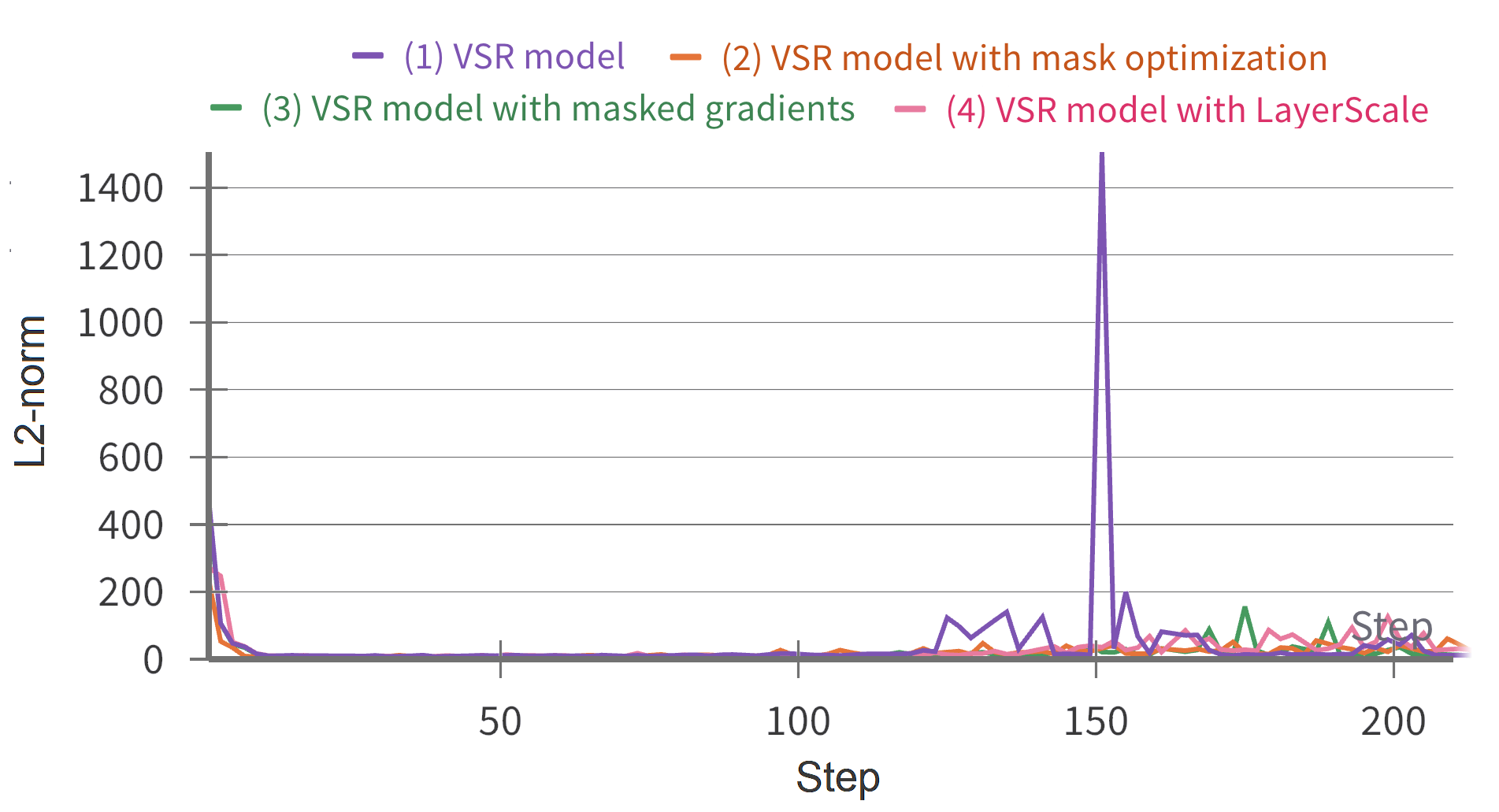}

To facilitate training VSR/AVSR models from scratch, we investigate their gradient norm when trained from scratch and observe a significant phenomenon of gradient explosion.
This phenomenon can be addressed by clipping the gradient to an appropriate value. However, as the network becomes deeper and more complex, the gradient also vanishes. Figure \ref{fig:gradient} illustrates this problem by showing the gradient norm of the first feed-forward layer in the VSR encoder. When no clipping is applied (grey line), the gradients explode, while only clipping (blue line) fails to solve vanishing problems. 
To address this issue, we investigate optimization strategies that enable the training of deeper high-capacity models. One approach is to optimize the Transformer/Conformer blocks, as suggested by various studies \cite{zhang2019fixup, bachlechner2021rezero, huang2020improving,touvron2021going}. Among these methods, LayerScale \cite{touvron2021going} stands out for regularizing deeper vision transformers. LayerScale incorporates a learnable diagonal matrix, initialized close to 0, to the output of each residual block to enhance the training dynamics. Another possible approach involves pruning weights to reduce the number of hidden units while maintaining the dense connectivity of the model \cite{zhu2017prune, mocanu2018scalable, evci2020rigging, yin2023dynamic}. For instance, the Gradual Magnitude Pruning (GMP) algorithm \cite{zhu2017prune} starts from a randomly initialized dense model and gradually prunes the smallest magnitude weights until it achieves the target sparsity. Encouragingly, their experiments demonstrate that large-sparse models consistently surpass small-dense models in various tasks with the same memory footprint and minimal performance loss.   

In this work, we introduce a mask optimization technique coined \textbf{MSRS} (Multimodal Speech Recognition from Scratch), which rapidly identifies a sparse topology within the dense model. This sparse topology receives a more stable gradient flow, and hence, stabilizes the training process, enabling training very large speech models without the need for any pretraining steps. Once the sparse mask becomes stable (within a few epochs), our approach allows for condensing to a dense model or maintaining a sparse model by only updating the non-zero values of the mask. Following, we outline our contributions as: (i) we present MSRS for efficient training of VSR/AVSR models from scratch; (ii) we investigate our approach in the limited-data regime and find that MSRS allows for training VSR models in situations where data is limited (around 90 hours). 
This makes it possible to extend to languages with scarce data available; (iii) we compare our method with numerous sparse training approaches, none of which can train large-scale VSR models from scratch without a pretrained frontend. We study the topological differences of their subnetworks and provide insight into why MSRS converges. We find that MSRS explicitly masks the weights that experience vanishing gradients to enhance training, whereas other methods that decouple initialization from the training process do not; (iv) additionally, we delve into an investigation that draws connections to LayerScale. This analysis undercores the complementary nature of MSRS and LayerScale in improving the gradient flow of deep speech models by regulating the width and depth of the model, respectively; (v)
finally, the effectiveness of MSRS is backed up with strong empirical results on VSR and AVSR tasks. For instance, despite being trained from scratch, our condensed VSR model reports competitive results on the LRS3 benchmark, while reducing training time more than \textbf{2$\times$}. Similarly, our condensed AVSR model outperforms the current state-of-the-art (SoTA) \cite{ma2023auto} by more than 3\% WER on highly corrupted speech, without using pretraining stages and under lower 16-float precision. 

\textit{Please note that the primary goal of this work is not to achieve efficiency through sparsity, but rather to explore the under-explored benefits of irregular sparse patterns in enabling the training of VSR/AVSR models from scratch.} 
\vspace{-0.1cm}

%% file: figures/gradient.tex
\begin{figure}[tb]
  \centering
  \includegraphics[width=0.76\columnwidth, trim=0cm 5cm 2.5cm 0cm, clip]{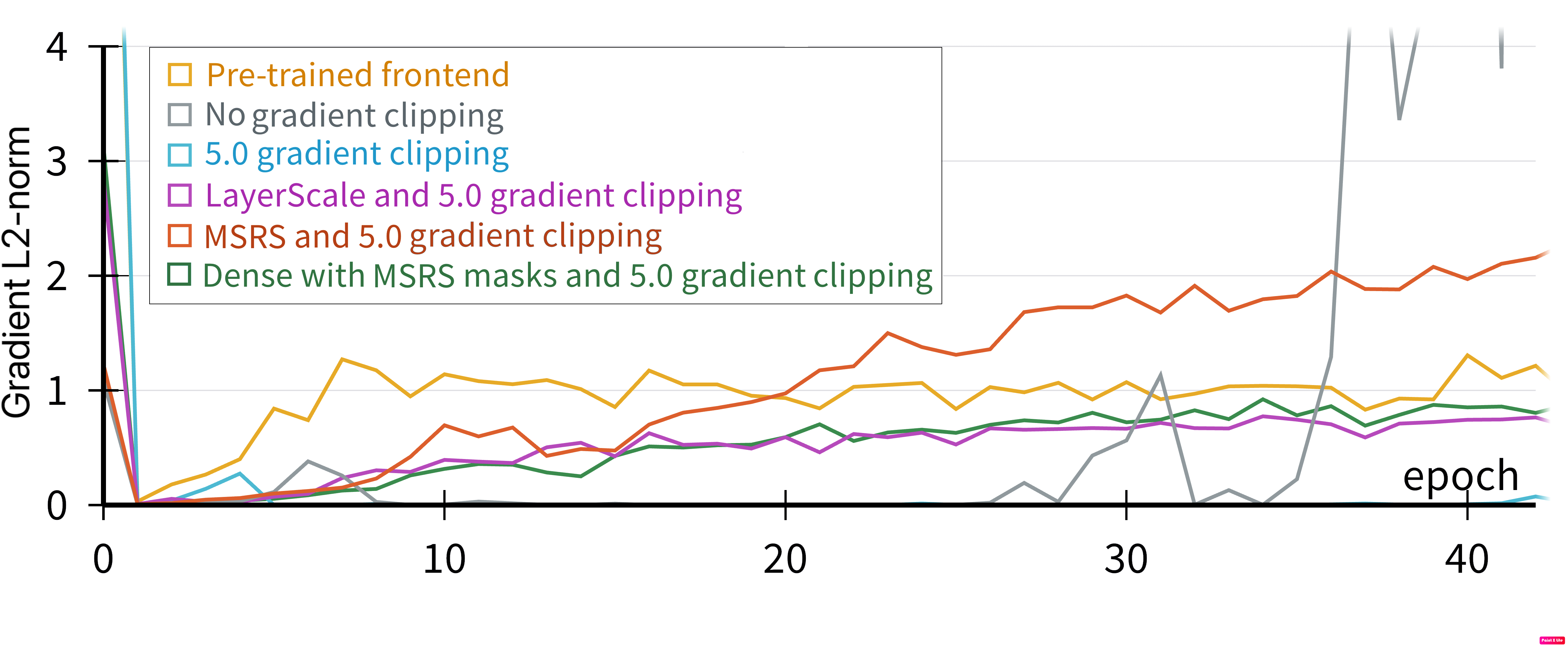}
  \vspace{-0.2cm}
  \captionsetup{font=scriptsize}
  \caption{A close-up view of half of VSR training process. L2 gradient norm of the first feed-forward layer in the encoder of multiple models trained with various regularization strategies. Assume uniform initializations, if no pre-training is specified.}% \honglie{I'd suggest to have a small title for each table and figure and highlight in bold, e.g., figure 1}}
  \vspace{-0.5cm}
  \label{fig:gradient}
\end{figure}
%: VSR model without gradient clipping, VSR model with gradient clipping at 5.0, VSR model with pretrained frontend and gradient clipping at 5.0, VSR model regularized with mask optimization and gradient clipping at 5.0, VSR model regularized with LayerScale and gradient clipping at 5.0. 

%% file: sections/method.tex
\section{Approach}
\vspace{-0.12cm}
\label{sec:approach}
For a speech recognition task, such as VSR, ASR, or AVSR, we have a model $f(\cdot; \theta)$ with model parameters $\theta$ that inputs a video, audio, or audio-visual sequence $x$ and outputs the corresponding transcription $y$ from a dataset $\mathcal{D} = \{(x_n, y_n)\}^{N}_{n=1}$ with $N$ sequences. This model is composed of a frontend for each modality, a multi-layer perceptron to combine the features from multiple domains (for AVSR), a Conformer encoder, and a Transformer decoder for joint training using connectionist temporal classification ($\mathcal{L}_{ctc}$) and attention ($\mathcal{L}_{att}$):
\vspace{-0.1cm}
\begin{align}
\label{eq:loss}
    \mathcal{L(\theta)} &= \mathcal{L}_{att} + \gamma \mathcal{L}_{ctc}
\end{align}
Consider a scenario where the model $f(\cdot; \theta)$ has millions parameters, making training complex and difficult to train from scratch \cite{shi2022learning, haliassos2022jointly}. To address this issue, we seek a mask $m\in\{0,1\}^p$ that can progressively shrink the width of the model $f(\cdot; \theta \odot m)$, resulting in a smaller and easily trainable model starting from the initial weights $\theta_0$, where $p$ represents the number of model parameters and $\odot$ is the element-wise product.
\subsection{Mask optimization}
\vspace{-0.12cm}
Inspired by \cite{gao2021adapting, colombo2021differentiable}, we simultaneously search for an optimal model structure $m\in\{0,1\}^p$ and its corresponding weights $\theta \in \mathbb{R}^p$, where $\theta$ is continuous in $\mathbb{R}$ and $m$ is discrete in $\{0,1\}$:
\vspace{-0.17cm}
\begin{align}
\label{eq:join_opt}
m_{*} &=  \argmin_{m}(\min_{\theta} \mathcal{L} (m, \theta), \mathcal{D}) \\
\label{eq:phi0}
\phi_0 &= \left (  \ln(\left | \theta_0 \right | + \varsigma )/2 + 1 \right ) \cdot \rho + \mu
\end{align}
This fact complicates optimization since $\mathcal{L}$ is not differentiable with respect to $m$, and standard gradient descent techniques cannot be directly applied. Therefore, we use a fully continuous approximation of that problem. Concretely, the sigmoid function is used to approximate the mask in continuous space: $m \rightarrow m_{l} = \sigma(l \phi)$, where $\phi_0$ is initialised near zero according to eq. (\ref{eq:phi0}) \ie~ small absolute values are mapped to negative and larger ones to positive. $l$ controls the \textit{sharpness} of the approximation, \ie~the larger, the closer to the binary function. The discrete optimization in eq. (\ref{eq:join_opt}) becomes continuous in eq. (\ref{eq:join_opt_cont}):
\vspace{-0.1cm}
\begin{align}
\label{eq:join_opt_cont}
& \phi_* = \argmin_{\phi} (\min_{\theta} \mathcal{L} (m_{l}, \theta), \mathcal{D}) \notag \\
\text{subject to \quad} &m_{l} = \sigma(l \phi) = \frac{1}{{1 + e^{-l \phi}}}, \quad l \gg 1 
\end{align}
In this scenario, $m_l$ is differentiable for all $l < \infty$, but when $\phi \neq 0$ and $l \rightarrow \infty$, $\nabla_{\phi}m_l = \sigma(l \phi) \cdot (1 - \sigma(l \phi)) \rightarrow 0$, which leads to vanishing gradients. 
To deal with that we consider using two-temperature minimization \cite{colombo2020disentangling}, \emph{i.e.,} use a larger value in the sigmoid during forward propagation $\scriptstyle \overrightarrow{l}$ and a smaller during backward propagation $\scriptstyle \overleftarrow{l}$. This results in gradient updates: % for $\phi$ and $\theta$:
\vspace{-0.47cm}
\begin{align}
\label{eq:updates_mask}
& \phi_{i+1} \leftarrow \phi_{i} - \eta_i  \nabla_{ m_{\scriptscriptstyle \overrightarrow{l}}} \mathcal{L}(z; m_{\scriptscriptstyle \overrightarrow{l}}, \theta) \nabla_{\phi}m_{\scriptscriptstyle \overleftarrow{l}} - \lambda\mathbf{1}_p \\
\label{eq:updates_model}
& \theta_{i+1} \leftarrow \theta_{i} - \alpha_i \nabla_{\theta} \mathcal{L}(z; m_{\scriptscriptstyle \overrightarrow{l}}, \theta)
\end{align}
% \honglie{Did we explain $\mathbf{1}$?}
where $\eta_i$ and $\alpha_i$ are learning rates, $z=(x, y)$ refers to input-output sequences, $\mathbf{1}_p = (1,1,...,1) \in \mathbb{R}^p$, and $\lambda$ is a hyperparameter that controls the pruning speed. The penalization term in eq. (\ref{eq:loss_mask}) encourages the entries of $\phi$ to be negative, leading to sparser masks.
\vspace{-0.1cm}
\begin{align}
\label{eq:loss_mask}
    &\mathcal{L}(m_l, \theta) = \mathcal{L}_{att} +  \gamma \mathcal{L}_{ctc} + \lambda {\mathbf{1}}_p^{\top} \phi
    \vspace{-0.2cm}
\end{align}
After $T$ epochs, a near-optimal binary mask can be obtained from $\phi_T$ by setting all positive values to 1 and all negative values to 0, where $\mathds{1}$ is the indicator function:
\vspace{-0.1cm}
\begin{equation}
\label{eq:mask}
m_* = \mathds{1} [\ \phi_T > 0]\ = \left\{\begin{matrix}
1 & \phi_T \geq   0\\ 
0 & \phi_T < 0
\end{matrix}\right.
\end{equation}
\subsection{Regularization early in training}
\vspace{-0.2cm}
\input{algorithms/approach}
Deep networks can have issues with vanishing or exploding gradients, which can make it difficult for them to converge effectively. This problem is particularly pronounced at the start of training, where regularization is essential for success. However, we have discovered that by using the appropriate $\lambda$ parameter, the mask gradually converges to a near-optimal sparsity early in the training process ($\sim$ 3 epochs) and remains stable for the remainder of the training. In fact, the mask learns to zero-out those weights that cause convergence issues, resulting in a narrower model that can benefit from depth. This is helpful because optimizing deep models and their masks at the same time throughout the entire training process can be computationally expensive. 
The algorithm for jointly optimizing the model and mask early in training is shown in Algorithm \ref{alg:mask_opt}. After that, we re-start the training on the regularized model using the generated mask and the model weights as initialization. At this step, we can choose whether to go back to a dense model or stay sparse by only updating the weights that correspond to the non-zero values of the mask. 
We reset the learning rate scheduler because models recover from pruning more easily with larger learning rates and struggle to recover with smaller ones \cite{liu2021sparse}.

%% file: algorithms/approach.tex
\vspace{-0.2cm}
\begin{algorithm}
\caption{Multimodal Speech Recognition from Scratch}
\label{alg:mask_opt}
\begin{algorithmic}
\State \textbf{Inputs:} model $f(\cdot, \theta_0, \phi_0)$; initial model $\theta_0$ and mask $\phi_0$ parameters; training data $\mathcal{D}$; hyperparameters $\scriptstyle \overrightarrow{l}$, $\scriptstyle \overleftarrow{l}$, $\epsilon$ and $\lambda$; learning rates $\eta$ and $\alpha$; number of epochs $T$; $dense$ flag.
\State Set step $j$=1
\While{\textit{stopping criteria} $< \epsilon$ or $j \leq T$} \Comment{\textit{stopping criteria} compares the sparsity between consecutive binary masks}
    \State Update $\phi_j$ following eq. (\ref{eq:updates_mask}) 
    \State Update $\theta_j$ following eq. (\ref{eq:updates_model}) 
    \State Increase step $j$   
\EndWhile
\State Get binary mask $m_{*}$ = $\mathds{1} [\ \phi_j > 0]\ $ and model weights $\theta_j$ 
\State Re-start $\eta$ and $\alpha$ and train from model weights $\theta_0=\theta_j \odot m_{*}$  
\For{$i$=1,...,$T$}
    \If{$dense$} \Comment{\textit{All weights are adjustable}} 
        \State Update $\theta_{i+1} \leftarrow \theta_{i} - \alpha_i \nabla_{\theta} \mathcal{L}(z; \mathbf{1}_p, \theta)$ 
    \Else \Comment{\textit{Only non-zero weights are adjustable}} 
        \State Update $\theta_{i+1} \leftarrow \theta_{i} - \alpha_i \nabla_{\theta} \mathcal{L}(z; m_{*}, \theta)$
    \EndIf
\EndFor
\State
\Return $\theta_T$, $m_{*}$
\end{algorithmic}
\end{algorithm}
\vspace{-0.2cm}

% \begin{algorithm}
% \caption{Mask optimization Early in Training - Mask opt.}
% \label{alg:mask_opt}
% \begin{algorithmic}
% \State \textbf{Inputs:} model $f(\cdot, \theta_0, \phi_0)$; initial model $\theta_0$ and mask $\phi_0$ parameters; training data $\mathcal{D}$; hyperparameters $l_f$, $l_b$, $\epsilon$ and $\gamma$; learning rates $\{\eta_i\}^T_{i=1}$ and $\{\beta_i\}^T_{i=1}$; number of training epochs $T$; $dense$ flag.
% \State
% \State $update$ $mask$ = \textit{True}
% \For{$j$=1,...,$T$}
%     \If{$update$ $mask$}
%         \State Update $\phi_i$ following eq. (\ref{eq:updates_mask})
%         \State Update $\theta_i$ following eq. (\ref{eq:updates_model})
%         \If{$stopping$ $criteria$ $< \epsilon$} \Comment{\textit{verify if the sparsity of consecutive binary masks is below $\epsilon$.}}
%             \State Set $update$ $mask$ = \textit{False}
%             \State Compute final binary mask $m*$ = $\mathds{1} [\ \phi_i > 0]\ $
%         \EndIf
%     \ElsIf{$dense$} \Comment{\textit{All weights are adjustable}} 
%         \State Update $\theta_{i+1} \leftarrow \theta_{i} - \beta_i \nabla_{\theta} \mathcal{L}(z; \mathbf{1}^p, \theta)$
%     \Else \Comment{\textit{only non-zero weights are adjustable}} 
%         \State Update $\theta_{i+1} \leftarrow \theta_{i} - \beta_i \nabla_{\theta} \mathcal{L}(z; m*, \theta)$
%     \EndIf
% \EndFor
% \State
% \Return $\theta_T$, $m*$
% \end{algorithmic}
% \end{algorithm}

%% file: sections/experimental_setup.tex
\vspace{-0.2cm}
\section{Experimental setup}
\label{sec:experimental_setup}
% \subsection{Dataset}
% \vspace{-0.2cm}
\par{\noindent \bf Dataset and evaluation.}
We use the LRS3 dataset \cite{afouras2018lrs3}, which is the largest publicly available audio-visual dataset in English and a standard benchmark for VSR/AVSR research. The dataset 
% was collected from TED and TEDx talks and 
contains 150,498 utterances for training (438 hours) and 1,321 utterances for testing (0.9 hours). Additionally, following previous works \cite{ma2023auto, fernandez2023sparsevsr}, we use VoxCeleb2 \cite{chung2018voxceleb2} and AVSpeech \cite{ephrat2018looking} audio-visual datasets for training, resulting in a total of 1,307 and 1,323 hours. We follow previous works~\cite{errattahi2018automatic} and report results using Word Error Rate (WER). 
% \vspace{-0.3cm}
% \subsection{Implementation details}
% \vspace{-0.2cm}
% \label{sec:setup}
\par{\noindent \bf Pre-processing.} We follow the pre-processing steps used in prior works \cite{ma2022visual, shi2022learning} to crop a 96$\times$96 region centred around the mouth and then transform each frame into a greyscale image. 
\par{\noindent \bf Data augmentation.} We apply horizontal flipping, random cropping, and adaptive time masking \cite{ma2022visual} to all our models. The masks we use are proportional to the length of the utterance and have a maximum masking length of up to 0.4 seconds.
\par{\noindent \bf VSR architecture details}. VSR baselines follow the architecture in \cite{fernandez2023sparsevsr}, which includes a small and a large model with 64 and 250 million parameters. Models consist of a 3D convolutional layer, a 2D ResNet-18 \cite{petridis2018end}, a Conformer encoder, a projection CTC layer, and a Transformer decoder. The small/large model has a 12-layer Conformer encoder with 256/768 inputs, 2,048/3,072 feed-forward dimensions, 4/12 attention heads, and a decoder with 6-layer Transformer with same dimensions and number of heads as the corresponding encoder. 
\par{\noindent \bf AVSR architecture details.} AVSR models are composed of a ResNet frontend for each modality \cite{ma2021towards}, a multi-layer perceptron for early fusion of multi-domain features, a single Conformer encoder, and a Transformer decoder. We use 2 models with 268/703 million parameters. The encoder consists of 12/20 layers with 768/1,024 inputs, 3,072/4,096 feed-forward dimensions, and 12/16 attention heads. The decoder is a 6/9-layer Transformer with the same dimensions and number of heads as the encoder. Audiovisual frontend outputs are concatenated and fed into a 2-layer MLP with 8,192 units and 768/1,024 outputs.
\par{\noindent \bf Training strategies.} The output is decoded using 5,000 unigram subwords. Two AdamW optimizers ($\beta_1=0.9$, $\beta_2=0.98$) are used with a cosine learning rate scheduler and gradient clipping at 5.0. The peak learning rates are set to 6e-4 for single dataset training and 1e-3 for multiple sets. The model is initially warmed up for up to 5 epochs of the $j$ epochs of joint training, and then warmed up again for 5 epochs during the subsequent 75 epochs of single training. Here, $j$ is the number of joint training epochs that are required for mask stabilization ($j$=3 if not otherwise specified). No language model is included. 
%If indicated, visual-only models incorporate a transformer language model (LM) trained with 166 million characters, as described in \cite{ma2023auto, ma2022visual}.
\par{\noindent \bf Mask hyper-parameters.} We set the initial mask weights $\phi_0$ to be very close to 0 with $\mu=$1e-3, $\rho=$5e-4 and $\varsigma=$1e-8. This allows both positive and negative weights to easily fluctuate at the beginning of training. We experimentally set up $\lambda$ = 2e-10, $\epsilon$ = 0.01, and hyperparameters $\scriptstyle \overrightarrow{l}$=$1e5$ to approximate the binary function and $\scriptstyle \overleftarrow{l}$=$1$ to avoid vanishing gradients.
\vspace{-0.2cm}

%% file: sections/results.tex
\section{Results}
\label{sec:resultsVSR}
\subsection{End-to-end training of VSR models from scratch}
\vspace{-0.15cm}
To compare the capability of VSR models trained from scratch, we conduct experiments on the LRS3 dataset and report the results in Table~\ref{tab:vsr_models}.
We observe that overall, both small and large dense models face convergence issues when the weights are randomly initialized. Comparing the small models trained from scratch, we show that using LayerScale~\cite{touvron2021going} or MSRS results in a WER from 100\,\% to 42.8\,\% and 48.5\,\%, respectively. The gain is likely due to a healthier gradient flow compared to the dense equivalent. Unlike LayerScale which adds extra parameters to the model, we further show that MSRS can benefit from sparse computation. For instance, when almost 40\% of the parameters are zeroed out, a gain of 1.2\,\% is observed. For large models, in contrast to previous works that solve the optimization issues by using pre-trained weights \cite{ma2023auto, ma2021towards, fernandez2023sparsevsr, haliassos2022jointly, shi2022learning}, MRSR results in a comparable WER of 21.1\,\%, while speeding up training by at least 2$\times$, depending on the pretraining strategy \cite{ma2022visual}. 
Similarly, when 40\% of the parameters are zeroed out, there is only an increase of 2.8\,\% in WER. This result is in line with our hypothesis that while resource-constrained settings may have limited capacity, large models, which are often challenging to train, can especially benefit from these regularization methods. Furthermore, a reduction of 0.7\,\% in WER is observed when combining both LayerScale and MSRS. This indicates that these two regularization strategies complement each other, where LayerScale almost eliminates the residual connections of the encoder at the start of training, limiting its depth, while MSRS quickly in training discovers sparse structures within the dense model, resulting in relatively smoother gradient flow compared with the dense counterpart.
\input{tables/VSR_models}

\subsection{Exploring the minimal amount of training data required for model convergence}
\vspace{-0.15cm}
To investigate the effect of data in this scenario, we trained our large VSR model with MSRS using varying amounts of training data and created sparse versions of it. As shown in Table \ref{tab:data_kept}, a sparse VSR model converges with an increase of up to a 5\% in WER when the data reduction reaches 97\%. This suggests the potential for training large models in situations where data is limited, such as for low resource languages. However, as the dataset continues to shrink, the pruning process slows down, which may require further adjustments to the $\lambda$ parameter. 
\input{tables/data_kept_table}
\subsection{Impact of the pruning speed on training from scratch}
\label{subsec:lambda}
\input{figures/lambda_analysis}
The success of training speech models from scratch using MSRS is highly dependent on the hyperparameter $\lambda$. An ablation study has been conducted to investigate the impact of different $\lambda$ values on the small VSR setup trained on LRS3, as shown in Figure \ref{fig:lambda_analysis}. The results indicate that when a good initialization (pretrained frontend) is used, $\lambda$ values within the range of 1e-12 to 1e-7 effectively prune the model to a near-optimal sparsity of around 40\%. Specifically, larger values of $\lambda$ result in faster pruning but also make recovery more difficult and lead to accuracy drops, while smaller values do not force enough weights to be negative, which will eventually result in a denser model. In the context of training from scratch, it is crucial to progressively eliminate weights that hinder the training process. We achieve this by zero-outing those weights that negatively impact the loss until a near-optimal sparsity is attained and a healthier gradient flow is established. Therefore, the target range of $\lambda$ becomes much smaller (1e-10 to 1e-9), with 2e-10 being the best trade-off in our all experiments. Note that if the dataset size decreases significantly below 300 hours (as shown in Table \ref{tab:data_kept}), it may be necessary to adjust the value of $\lambda$ to maintain the pruning speed. 

\vspace{-0.2cm}
\subsection{End-to-end training of AVSR models from scratch}
\vspace{-0.15cm}
Table \ref{tab:AVSR} presents the AVSR results that illustrate the capability of other input-based speech models to be trained from scratch using MSRS. Even though, our proposed AVSR model with small capacity can be fully trained from scratch without additional regularization, it is crucial to use regularization as the model size increases to ensure convergence.  Therefore, we explore multiple capabilities of MSRS and utilize a very large model (700 M parameters) that benefits from depth. Specifically, we leverage MSRS to require minimal data for convergence and accelerate training by using less than 1k hours of the available data (only 30\%) for MSRS training with 16-float precision, and then transitioned to a dense model trained on the entire dataset to achieve SoTA results. Our very large model outperforms the current SoTA method presented in \cite{ma2023auto} without using any pretraining strategies under lower bit precision. This is particularly evident under very noisy conditions, where we observe more than a 3\% WER absolute improvement. This is not the case for LayerScale, where a lower precision model does not converge. 
\input{tables/AVSR_models}
\vspace{-0.3cm}
\subsection{Comparison with other sparse-based training methods}
\vspace{-0.15cm}
\label{subsec:ablation}
\input{tables/ablation_pretrainedmodels_small}

We compare various pruning methods for neural networks in Table \ref{tab:ablation_pretrained}, focusing on those that achieve the desired sparsity during or at the start of training. One method is GMP \cite{zhu2017prune}, which gradually prunes the smallest magnitude weights during training to reach the target sparsity. Another approach is sparse-to-sparse training, which starts with the target sparsity and uses a prune-and-grow approach to explore the parameter space. Notable methods in this category include SET \cite{mocanu2018scalable}, RigL \cite{evci2020rigging} and CHASE \cite{yin2023dynamic}, which use Erdos-Rényi-Kernel (ERK) \cite{mocanu2018scalable, evci2020rigging} to initialize models at the target sparsity and then prune weights based on magnitude or gradient. Our results show that when a pretrained frontend is given, all sparse strategies work without much degradation, but when the model is randomly initialized, only MSRS converges. 
The main difference between MSRS and other methods is that MSRS selectively prunes weights that negatively impact the loss, resulting in a narrower model that enables proper gradient flow. In contrast, other sparse methods employ initialization and pruning strategies that are decoupled from the training process. Figure \ref{fig:sparsity_layers} shows the sparsity for each module using different pruning methods. We observe that MSRS prunes frontend and encoder layers significantly to improve gradient flow, while ERK initialization primarily targets layers with more weights in the frontend, which results in less healthy flow. Our insight is that MSRS solves the vanishing gradient issue by properly reducing model complexity. This is achieved by cutting down small gradients, which are the root cause of the issue. To prove that, we applied the generated masks from MSRS training to the gradients of a dense model during training and found improved convergence (Fig. \ref{fig:gradient} - green line), with 52.1\% WER, only 5\% up. This implies that the masked weights are those that experience gradient vanishing.
%support the idea that SRS reduces model complexity. %, which enables training from scratch.
\input{figures/sparsity_layers}

%% file: tables/VSR_models.tex
\begin{table}[tb]
\centering
\setlength{\tabcolsep}{2.5pt}
\captionsetup{font=scriptsize}
\caption{VSR models under different initializations ($\theta_0$) and regularizations on the LRS3 test set. The training data $\mathcal{D}$ is LRS3 (S) or a combination of LRS3, VoxCeleb2 and AVSpeech (L). S contains 438 hours, L contains 3,068 hours, L$^*$ contains 1,759 hours, L$^\dagger$ contains 3,448 hours and L$^\ddagger$ contains 100k private hours.}%\shiwei{is it possible to add a column to indicate weather the method is trainable or not? same for the following tables}}%\shiwei{bold results usually are the best results not our results...}}% \shiwei{It is preferable to highlight our results in bold.}}
\vspace{-0.3cm}
\resizebox{0.9\columnwidth}{!}{
\begin{tabular}{lcccccc}
\toprule
\textbf{Methods} & \textbf{$\theta_0$} & \textbf{$\#$Params} & \textbf{Sparsity} & \textbf{$\mathcal{D}$} & \textbf{WER (\%)} & \textbf{Trainable}  \\ \midrule
% SparseVSR \cite{fernandez2023sparsevsr}                   & Frontend                      & 56 M               & 0               & S             & 39.3           \\ \hdashline
\textit{Small models} \\ \midrule
Dense                   & Random                        & 56 M               & 0               & S             & 100.0 & \xmark \\\hdashline[1pt/1pt] 
LayerScale              & Random                        & 56 M               & 0               & S             & \textbf{42.8}  & \cmark       \\
MSRS             & Random                        & 56 M               & 0               & S             & 48.5   & \cmark          \\
MSRS             & Random                        & 32 M               & 43.8           & S             & 47.3 & \cmark        \\ \midrule
\textit{Large models} \\ \midrule
Dense                   & Random                        & 250 M              & 0               & S/L         & 100.0  & \xmark \\
AVHubert \cite{shi2022learning} & Random & 325 M & 0 & S & 62.3 & \xmark \\
RAVEn \cite{haliassos2022jointly} & Random & 493 M& 0 & S & 87.3 & \xmark \\ 
MSRS & Random & 250 M& 39.1 & S & \textbf{46.9} & \cmark \\ \hdashline[1pt/1pt]
AVHubert \cite{shi2022learning} & Encoder & 325 M & 0 & $L^*$ & 26.9 & \cmark  \\
RAVEn \cite{haliassos2022jointly} & Encoder & 493 M& 0 & $L^*$ &24.4 & \cmark \\
SparseVSR \cite{fernandez2023sparsevsr}                   & Frontend                      & 250 M     & 0               & L         & 21.1     & \cmark       \\         
Auto-AVSR\cite{ma2023auto} & Frontend & 250 M & 0 & $L^\dagger$ & 20.5 & \cmark \\ 
LP Conformer \cite{chang2024conformer} & Encoder & 0.57 B & 0 & $L^\ddagger$ & \textbf{12.8} & \cmark \\ 
\hdashline[1pt/1pt]
LayerScale              & Random                        & 250 M              & 0               & L         & 21.9     & \cmark     \\
MSRS             & Random                        & 250 M               & 0               & L         & \textbf{21.1}   & \cmark     \\
MSRS             & Random                        & 151 M              & 39.7            & L         & 23.9  & \cmark \\
MSRS + LayerScale & Random & 151 M & 39.7 & L & \textbf{23.2} & \cmark \\
\bottomrule
\end{tabular}
}
%\small\textsuperscript{{$^*$} LRS3 + VoxCeleb2 (1,759 hours)} \\ \small\textsuperscript{{$^\dagger$} LRW + LRS2 + LRS3 + VoxCeleb2 + AVSpeech (3,448 hours).}
  \vspace{-1.8em}
\label{tab:vsr_models}
\end{table}

%% file: tables/data_kept_table.tex
% \begin{table}[htb]
% \setlength{\tabcolsep}{1.5pt}
% \renewcommand{\arraystretch}{1}
% \caption{Sparse large VSR model trained with SRS on varying amounts of data from a combination of the LRS3, VoxCeleb2, and AVSpeech datasets. Evaluation on the LRS3 test set.}
% % \honglie{Do we know how the pretrained model performs with 10 \% data}}
% \vspace{-0.2cm}
% \label{tab:data_kept}
% \centering
% \resizebox{\columnwidth}{!}{
% \begin{tabular}{lccccccccc}
% \toprule
% \textbf{Data kept (\%)} & 100  & 75   & 50   & 25   & 10  & 5 & 3 & 1\\ \midrule
% \textbf{Data in hours} & 3,068  & 2,301   & 1,534   & 767   & 307   & 153 & 92 & 31\\ \midrule
% \textbf{Epochs ($j+T$) }       & $3+75$ &$3+75$ & $3+75$ & $3+150$  & $5+300$ & $10+600$ & $15+800$ & $45+950$\\ \midrule
% \textbf{Sparsity (\%)}  & 39.7 & 39.2 & 39.2 & 39.1  & 39.1  & 39.1 & 39.2 & 39.1\\ \midrule
% \textbf{WER (\%) }      & 23.9 & 24.1 & 24.2 & 25.3 & 26.7 & 26.9 & 28.5 & 37.7\\ \bottomrule
% \end{tabular}}
%   \vspace{-1.3em}
% \end{table}
\begin{table}[htb]
\setlength{\tabcolsep}{1.5pt}
\renewcommand{\arraystretch}{1}
\captionsetup{font=scriptsize}
\caption{Sparse large VSR model trained with MSRS on varying amounts of data from a combination of the LRS3, VoxCeleb2, and AVSpeech datasets. Evaluation on the LRS3 test set.}
% \honglie{Do we know how the pretrained model performs with 10 \% data}}
\vspace{-0.2cm}
\label{tab:data_kept}
\centering
\resizebox{0.95\columnwidth}{!}{
\begin{tabular}{ccrclccc}
\toprule
\textbf{Data (\%)} & \textbf{Data (hrs)} & \multicolumn{3}{c}{\textbf{Epochs (j + T)}} & \textbf{Sparsity} & \textbf{WER (\%)} & \textbf{Trainable} \\ \midrule
100                     & 3,068                      & $3$& $+$&$ 75$                               & 39.7              & 23.9              &  \cmark                  \\
75                      & 2,301                      & $3$& $+$&$ 75$                              & 39.2              & 24.1              &    \cmark                \\
50                      & 1,534                      & $3$& $+$&$ 75$                                & 39.2              & 24.2              &  \cmark                  \\
25                      & 767                        & $3$& $ +$&$ 150$                               & 39.1              & 25.3              &  \cmark                  \\
10                      & 307                        & $5$& $ +$&$ 300$                              & 39.1              & 26.7              &  \cmark                  \\
5                       & 153                        & $10$& $ +$&$ 600$                               & 39.1              & 26.9              &  \cmark                  \\
3                       & 92                         & $15$& $ +$&$ 800$                              & 39.2              & 28.5              &  \cmark                  \\
1                       & 31                         & $45$& $ +$&$ 950$                               & 39.1              & 37.7              &   \cmark               \\ \bottomrule 
\end{tabular}}
  \vspace{-1em}
\end{table}

%Data 10%, WER 25.4, 20.3% sparsity, 300 epochs

%% file: figures/lambda_analysis.tex
\vspace{-0.5em}
\begin{figure}[t]
  \centering

  \includegraphics[width=0.83\linewidth, trim=0.5cm 0.5cm 0.4cm 0cm, clip]{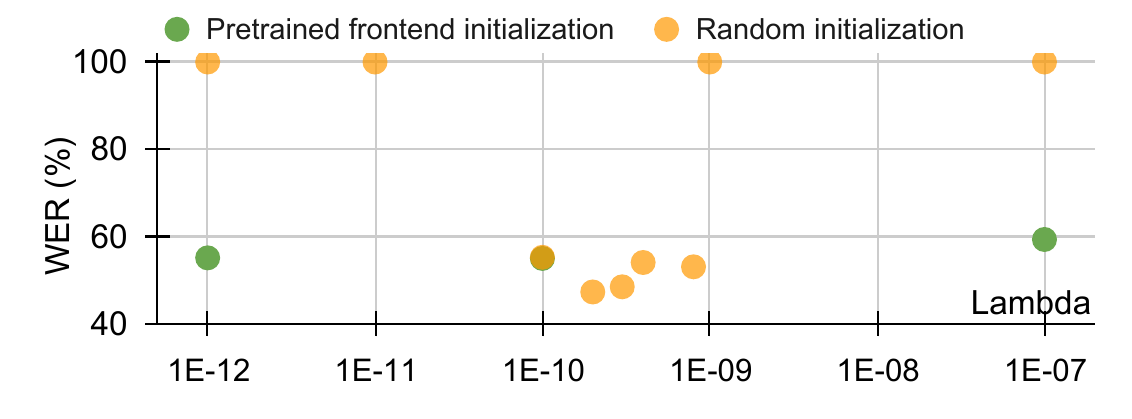}
  \vspace{-0.2cm}
  \captionsetup{font=scriptsize}
  \caption{Impact of $\lambda$ on VSR trainings.}
  \vspace{-2em}
  \label{fig:lambda_analysis}
  
\end{figure}

%% file: tables/AVSR_models.tex
\vspace{-1em}
\begin{table}[b]
\setlength{\tabcolsep}{1.6pt}
\vspace{-0.4cm}
\captionsetup{font=scriptsize}
\caption{WER (\%) of AVSR models as a function of the noise levels on the LRS3 test set. Models are trained using a combination of LRS3, VoxCeleb2, and AVSpeech datasets in the presence of babble noise$^*$. ``Reg.'' refers to regularization.}
\vspace{-0.3cm}
\centering
\resizebox{0.92\columnwidth}{!}{
\begin{tabular}{lccccccccc}
\toprule
\textbf{Reg. / SNR (dB)} & \textbf{$\#$Params} & \textbf{-7.5} & \textbf{-2.5} & \textbf{2.5} & \textbf{7.5} & \textbf{12.5} & \textbf{Clean} & \textbf{Trainable} \\ \midrule
Auto-AVSR \cite{ma2023auto}                       & 443 M                                  & 5.6           & 2.2           & 1.5          & 1.0          & 1.0           & 0.9   & \cmark          \\ \midrule
Dense                            & 268 M                                  & 2.8          & 1.5          & 1.1         & 1.0         & 0.9 & \textbf{0.8}    & \cmark        \\
Dense MSRS                        & 268 M                                  & 2.6           & 1.5           & 1.1 & 1.0 & 0.9           & 0.9 & \cmark   \\
Sparse MSRS (39.7\%)                      & 162 M                              & 3.6           & 1.8           & 1.5          & 1.1          & 1.0           & 1.0      & \cmark       \\\midrule
Dense MSRS$^\dagger$                        & 703 M                                  & \textbf{2.5}           & \textbf{1.3}           & \textbf{1.0}          & \textbf{0.9}          & \textbf{0.9}           & 0.9       & \cmark     \\ \bottomrule
%SRS$^\dagger$                        & 703 M      & 0                            & 2.7           & 1.4           & 1.0          & 0.9          & 0.9           & 0.8           \\ \bottomrule
% SRS$^\dagger$                        & 275 M      & 39.1                            &  3.4          &     1.7       &    1.1       &   1.1        &   0.9         &   0.9         \\ \bottomrule
\end{tabular}
}
\label{tab:AVSR}
\small\textsuperscript{$^*$Training/testing data augmented with babble noise from NOISEX \cite{varga1993assessment}. $^\dagger$Precision FP-16.} 
  \vspace{-1.8em}
\end{table}

%% file: tables/ablation_pretrainedmodels_small.tex
% \begin{table}[tb]
% \caption{VSR models sparsified with different pruning strategies. Models follow different initializations $\theta_0$, where $p$ is the number of model parameters, and $s$ is the sparsity. We use LRS3 for training and test. }
% \setlength{\tabcolsep}{3pt}
% \centering
% \begin{tabular}{lcrrr}
% \toprule
% \textbf{Regularization} & \textbf{$\theta_0$} & $p$ & $s$     & \textbf{WER}    \\ \midrule
% Dense          & Frontend   & 56 M      & 0\%    & 39.3\% \\
% LRR \cite{fernandez2023sparsevsr}          & Model      & 56 M      & 40.0\% & 38.5\% \\
% Mask opt.        & Model      & 56 M      & 41.7\% & 39.0\% \\ \midrule
% SET \cite{mocanu2018scalable}            & Frontend   & 56 M      & 40.0\% & 44.6\% \\
% RigL \cite{evci2020rigging}          & Frontend   & 56 M      & 40.0\% & 44.4\% \\
% GMP  \cite{zhu2017prune}          & Frontend   & 56 M      & 40.0\% & 42.7\% \\
% CHASE  \cite{yin2023dynamic}        & Frontend   & 56 M      & 40.0\% & 42.7\% \\
% Mask opt.        & Frontend   & 56 M      & 41.7\% & 42.5\%   \\ \midrule
% Dense          & Random     & 56 M      & 0\%    & 100.0\%  \\
% SET \cite{mocanu2018scalable}           & Random     & 56 M      & 40.0\% & 100.0\%  \\
% RigL \cite{evci2020rigging}          & Random     & 56 M      & 40.0\% & 100.0\%  \\
% GMP  \cite{zhu2017prune}          & Random     & 56 M      & 40.0\% & 100.0\%  \\
% CHASE \cite{yin2023dynamic}         & Random     & 56 M      & 40.0\% & 100.0\%  \\
% Mask opt.       & Random     & 56 M      & 43.8\% & 47.3\% \\
% \bottomrule
% \end{tabular}
% \label{tab:ablation_pretrained}
% \end{table}
\begin{table}[tb]
\captionsetup{font=scriptsize}
\caption{VSR models sparsified with different strategies on the LRS3 test set. Models follow different initializations $\theta_0$.}
\vspace{-0.3cm}
\setlength{\tabcolsep}{2.5pt}
\centering
\resizebox{0.92\columnwidth}{!}{
\begin{tabular}{lccccc}
\toprule
\textbf{Methods} & \textbf{$\theta_0$} & \textbf{$\#$Params} & \multicolumn{1}{c}{\textbf{Sparsity}}     & \textbf{WER (\%)}  & \textbf{Trainable}  \\ \midrule
SparseVSR \cite{fernandez2023sparsevsr}          & Model      & 34 M      & 40.0 & 38.5 & \cmark \\
MSRS        & Model      & 33 M      & 41.7 & 39.0 & \cmark \\ \midrule
SparseVSR \cite{fernandez2023sparsevsr}          & Frontend   & 56 M      & 0    & 39.3 & \cmark \\
SET \cite{mocanu2018scalable}            & Frontend   & 34 M      & 40.0 & 44.6 & \cmark \\
RigL \cite{evci2020rigging}          & Frontend   & 34 M      & 40.0 & 44.4 & \cmark \\
GMP  \cite{zhu2017prune}          & Frontend   & 34 M      & 40.0 & 44.7 & \cmark \\
CHASE  \cite{yin2023dynamic}        & Frontend   & 34 M      & 40.0 & 42.7 & \cmark \\
MSRS        & Frontend   & 33 M      & 41.7 & \textbf{42.5}   & \cmark \\ \midrule
Dense          & Random     & 56 M      & 0    & 100.0 & \xmark  \\
SET \cite{mocanu2018scalable}, RigL \cite{evci2020rigging}           & \multirow{2}{*}{Random}     & \multirow{2}{*}{34 M}      & \multirow{2}{*}{40.0} & \multirow{2}{*}{100.0}  & \xmark \\
GMP  \cite{zhu2017prune}, CHASE \cite{yin2023dynamic}          &      &       &  &  & \\
MSRS       & Random     & 32 M      & 43.8 & \textbf{47.3} & \cmark \\ \midrule
Dense with MSRS masks      & Random     & 56 M      & 0 & 52.1 & \cmark \\
\bottomrule
\end{tabular}
}
\vspace{-0.6em}
\label{tab:ablation_pretrained}
\end{table}

%% file: figures/sparsity_layers.tex
\begin{figure}[t]
  \centering
  \includegraphics[width=0.95\columnwidth,trim=0.5cm 2.1cm 2.6cm 0.7cm, clip]{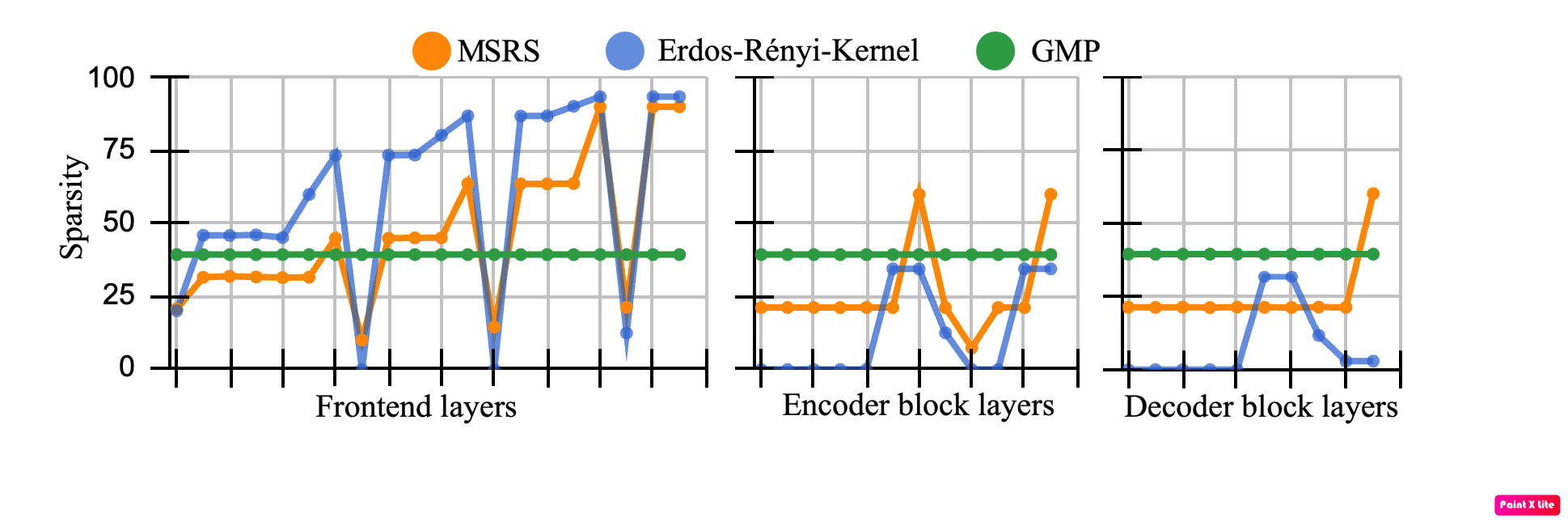}
  \vspace{-0.25cm}
  \captionsetup{font=scriptsize}
  \caption{Sparsity distribution in the frontend, an encoder block and a decoder block using various approaches: MSRS, ERK (for SET, RigL, and CHASE) and GMP. For encoder and decoder, the average/block is presented, respectively. Non-prunable layers like normalization layers are excluded.}
  \label{fig:sparsity_layers}
    \vspace{-1.8em}
\end{figure}

%% file: sections/conclusions.tex
\vspace{-0.6cm}
\section{Conclusions}
\label{sec:conclusion}
\vspace{-0.1cm}
In this work, we introduce a sparse regularization approach that focuses on quickly developing sparse patterns within dense models to promote gradient flow, enabling end-to-end training of VSR/AVSR models from scratch. Our algorithm efficiently obtains a nearly optimal mask in just a few epochs and the models can be transferred back to dense or kept sparse until completion. Our experiments show competitive results in both VSR and AVSR modalities, with AVSR particularly standing out for its improvements in noisy scenarios. MSRS is the only sparse technique that addresses the convergence issues faced by dense models when training from scratch, by selectively masking the weights affected by vanishing gradients. Moreover, it allows lower bit precision without adding extra parameters.